\documentclass{article}
\usepackage{ijcai22}

\usepackage{times}
\usepackage{soul}
\usepackage{url}
\usepackage[utf8]{inputenc}
\usepackage[small]{caption}
\usepackage{graphicx}
\usepackage{amsmath}
\usepackage{amsthm,amssymb}
\usepackage{booktabs}
\usepackage{algorithm}
\usepackage{algorithmic}
\usepackage{epstopdf}
\usepackage{tikz}
\usepackage{xcolor}
\usepackage{pgfplots}
\usepackage{booktabs}
\usetikzlibrary{spy}
\pgfplotsset{width=11cm,height=8cm}
\usepackage{overpic}
\usepackage{etex}
\usepackage{tikz}
\usepackage{pgfplots}
\usetikzlibrary{spy,calc}
\usepackage{array}
\newcolumntype{x}[1]{>{\centering\arraybackslash\hspace{0pt}}p{#1}}
\definecolor{sh_gray}{rgb}{0.84,0.84,0.84}
\definecolor{sh_gray2}{rgb}{1,0.89,0.75}
\definecolor{color3}{rgb}{0.95,0.95,0.95}
\definecolor{color4}{rgb}{0.96,0.96,0.86}
\definecolor{color5}{rgb}{0.90,0.90,0.90}

\newlength{\Oldarrayrulewidth}

\newif\ifblackandwhitecycle
\gdef\patternnumber{0}

\pgfkeys{/tikz/.cd,
    zoombox paths/.style={
        draw=orange,
        very thick
    },
    black and white/.is choice,
    black and white/.default=static,
    black and white/static/.style={
        draw=white,
        zoombox paths/.append style={
            draw=white,
            postaction={
                draw=black,
                loosely dashed
            }
        }
    },
    black and white/static/.code={
        \gdef\patternnumber{1}
    },
    black and white/cycle/.code={
        \blackandwhitecycletrue
        \gdef\patternnumber{1}
    },
    black and white pattern/.is choice,
    black and white pattern/0/.style={},
    black and white pattern/1/.style={
            draw=white,
            postaction={
                draw=black,
                dash pattern=on 2pt off 2pt
            }
    },
    black and white pattern/2/.style={
            draw=white,
            postaction={
                draw=black,
                dash pattern=on 4pt off 4pt
            }
    },
    black and white pattern/3/.style={
            draw=white,
            postaction={
                draw=black,
                dash pattern=on 4pt off 4pt on 1pt off 4pt
            }
    },
    black and white pattern/4/.style={
            draw=white,
            postaction={
                draw=black,
                dash pattern=on 4pt off 2pt on 2 pt off 2pt on 2 pt off 2pt
            }
    },
    zoomboxarray inner gap/.initial=5pt,
    zoomboxarray columns/.initial=2,
    zoomboxarray rows/.initial=2,
    subfigurename/.initial={},
    figurename/.initial={zoombox},
    zoomboxarray/.style={
        execute at begin picture={
            \begin{scope}[
                spy using outlines={%
                    zoombox paths,
                    width=\imagewidth / \pgfkeysvalueof{/tikz/zoomboxarray columns} - (\pgfkeysvalueof{/tikz/zoomboxarray columns} - 1) / \pgfkeysvalueof{/tikz/zoomboxarray columns} * \pgfkeysvalueof{/tikz/zoomboxarray inner gap} -\pgflinewidth,
                    height=\imageheight / \pgfkeysvalueof{/tikz/zoomboxarray rows} - (\pgfkeysvalueof{/tikz/zoomboxarray rows} - 1) / \pgfkeysvalueof{/tikz/zoomboxarray rows} * \pgfkeysvalueof{/tikz/zoomboxarray inner gap}-\pgflinewidth,
                    magnification=3,
                    every spy on node/.style={
                        zoombox paths
                    },
                    every spy in node/.style={
                        zoombox paths
                    }
                }
            ]
        },
        execute at end picture={
            \end{scope}
     \gdef\patternnumber{0}
        },
        spymargin/.initial=0.5em,
        zoomboxes xshift/.initial=1,
        zoomboxes right/.code=\pgfkeys{/tikz/zoomboxes xshift=1},
        zoomboxes left/.code=\pgfkeys{/tikz/zoomboxes xshift=-1},
        zoomboxes yshift/.initial=0,
        zoomboxes above/.code={
            \pgfkeys{/tikz/zoomboxes yshift=1},
            \pgfkeys{/tikz/zoomboxes xshift=0}
        },
        zoomboxes below/.code={
            \pgfkeys{/tikz/zoomboxes yshift=-1},
            \pgfkeys{/tikz/zoomboxes xshift=0}
        },
        caption margin/.initial=0.4ex,
    },
    adjust caption spacing/.code={},
    image container/.style={
        inner sep=0pt,
        at=(image.north),
        anchor=north,
        adjust caption spacing
    },
    zoomboxes container/.style={
        inner sep=0pt,
        at=(image.north),
        anchor=north,
        name=zoomboxes container,
        xshift=\pgfkeysvalueof{/tikz/zoomboxes xshift}*(\imagewidth+\pgfkeysvalueof{/tikz/spymargin}),
        yshift=\pgfkeysvalueof{/tikz/zoomboxes yshift}*(\imageheight+\pgfkeysvalueof{/tikz/spymargin}+\pgfkeysvalueof{/tikz/caption margin}),
        adjust caption spacing
    },
    calculate dimensions/.code={
        \pgfpointdiff{\pgfpointanchor{image}{south west} }{\pgfpointanchor{image}{north east} }
        \pgfgetlastxy{\imagewidth}{\imageheight}
        \global\let\imagewidth=\imagewidth
        \global\let\imageheight=\imageheight
        \gdef\columncount{1}
        \gdef\rowcount{1}
        
    },
    image node/.style={
        inner sep=0pt,
        name=image,
        anchor=south west,
        append after command={
            [calculate dimensions]
            node [image container,subfigurename=\pgfkeysvalueof{/tikz/figurename}-image] {\phantomimage}
            node [zoomboxes container,subfigurename=\pgfkeysvalueof{/tikz/figurename}-zoom] {\phantomimage}
        }
    },
    color code/.style={
        zoombox paths/.append style={draw=#1}
    },
    connect zoomboxes/.style={
    spy connection path={\draw[draw=none,zoombox paths] (tikzspyonnode) -- (tikzspyinnode);}
    },
    help grid code/.code={
        \begin{scope}[
                x={(image.south east)},
                y={(image.north west)},
                font=\footnotesize,
                help lines,
                overlay
            ]
            \foreach \x in {0,1,...,9} {
                \draw(\x/10,0) -- (\x/10,1);
                \node [anchor=north] at (\x/10,0) {0.\x};
            }
            \foreach \y in {0,1,...,9} {
                \draw(0,\y/10) -- (1,\y/10);                        \node [anchor=east] at (0,\y/10) {0.\y};
            }
        \end{scope}
    },
    help grid/.style={
        append after command={
            [help grid code]
        }
    },
}

\newcommand\phantomimage{%
    \phantom{%
        \rule{\imagewidth}{\imageheight}%
    }%
}
\newcommand\zoombox[2][]{
    \begin{scope}[zoombox paths]
        \pgfmathsetmacro\xpos{
            (\columncount-1)*(\imagewidth / \pgfkeysvalueof{/tikz/zoomboxarray columns} + \pgfkeysvalueof{/tikz/zoomboxarray inner gap} / \pgfkeysvalueof{/tikz/zoomboxarray columns} ) + \pgflinewidth
        }
        \pgfmathsetmacro\ypos{
            (\rowcount-1)*( \imageheight / \pgfkeysvalueof{/tikz/zoomboxarray rows} + \pgfkeysvalueof{/tikz/zoomboxarray inner gap} / \pgfkeysvalueof{/tikz/zoomboxarray rows} ) + 0.5*\pgflinewidth
        }
        \edef\dospy{\noexpand\spy [
            #1,
            zoombox paths/.append style={
                black and white pattern=\patternnumber
            },
            every spy on node/.append style={#1},
            x=\imagewidth,
            y=\imageheight
        ] on (#2) in node [anchor=north west] at ($(zoomboxes container.north west)+(\xpos pt,-\ypos pt)$);}
        \dospy
        \pgfmathtruncatemacro\pgfmathresult{ifthenelse(\columncount==\pgfkeysvalueof{/tikz/zoomboxarray columns},\rowcount+1,\rowcount)}
        \global\let\rowcount=\pgfmathresult
        \pgfmathtruncatemacro\pgfmathresult{ifthenelse(\columncount==\pgfkeysvalueof{/tikz/zoomboxarray columns},1,\columncount+1)}
        \global\let\columncount=\pgfmathresult
        \ifblackandwhitecycle
            \pgfmathtruncatemacro{\newpatternnumber}{\patternnumber+1}
            \global\edef\patternnumber{\newpatternnumber}
        \fi
    \end{scope}
}

\title{Linear Array Network for Low-light Image Enhancement}

\author{
Keqi Wang$^1$
\and
Ziteng Cui$^{2,3}$\and
Jieru Jia$^{1}$\and
Hao Xu$^{1}$\and
Ge Wu$^{5}$\and
Yin Zhuang$^{4}$\and\\
Lu Chen$^{1}$\and
Zhiguo Hu$^{1}$\And
Yuhua Qian$^{1}$
    \affiliations
    $^1$ShanXi University,
    $^2$Shanghai Jiao Tong University,
    $^3$Shanghai AI Lab,\\
    $^4$Megvii Inc.,
    $^5$Beijing Institute of Technology\\
    \emails
    wkq\_2021@163.com
}
\begin{document}

\maketitle

\begin{abstract}
Convolution neural networks (CNNs) based methods have dominated the low-light image enhancement tasks due to their outstanding performance.
However, the convolution operation is based on a local sliding window mechanism, which is difficult to construct the long-range dependencies of the feature maps. Meanwhile, the self-attention based global relationship aggregation methods have been widely used in computer vision, but these methods are difficult to handle high-resolution images because of the high computational complexity.
To solve this problem, this paper proposes a Linear Array Self-attention (LASA) mechanism, which uses only two 2-D feature encodings to construct 3-D global weights and then refines feature maps generated by convolution layers.
Based on LASA, Linear Array Network (LAN) is proposed, which is superior to the existing state-of-the-art (SOTA) methods in both RGB and RAW based low-light enhancement tasks with a smaller amount of parameters.
%We will release the source code at Github upon the acceptance of this submission.
The code is released in \url{https://github.com/cuiziteng/LASA_enhancement}.
\end{abstract}

\section{Introduction}

% Imaging system always plays a fundamental and critical role either for modern artificial intelligence or for human daily demand.
% However, images captured under sub-optimal lighting environments will appear different kinds of degradations, such as low contrast, color cast, and intensive noise, etc.
% These images not only affect human aesthetic quality, but also unsatisfactory demand for high-level computer vision tasks.
% As imaging is a irreversible processing, image enhancement methods are a effectiveness means to improve images quality.

Camera imaging system plays a fundamental role in modern artificial intelligence (AI) applications and human's daily demand. However, images captured under low-light condition suffer from different kinds of degradation, such as undesired color distortion and in-camera noises. This would give an negative effect on human visual experience and downstream vision tasks, such as
%semantic segmentation~\cite{DANNet_CVPR21} and
object detection
~\cite{ICCV_MAET}, visual recognition~\cite{mao2021dual}
and et al.
%Low-light image mainly has two sub-directions
There are mainly two sub-directions for recovering low-light images: one is mapping low-light RGB image to its' normal-light RGB counterpart~\cite{LLNet,RetinexNet},
while another is to map low-light RAW measurements to normal-light RGB counterpart as an image signal processing (ISP) process~\cite{see_in_the_dark}.

\begin{figure}[htb]
\centering
\includegraphics[width=2.7in]{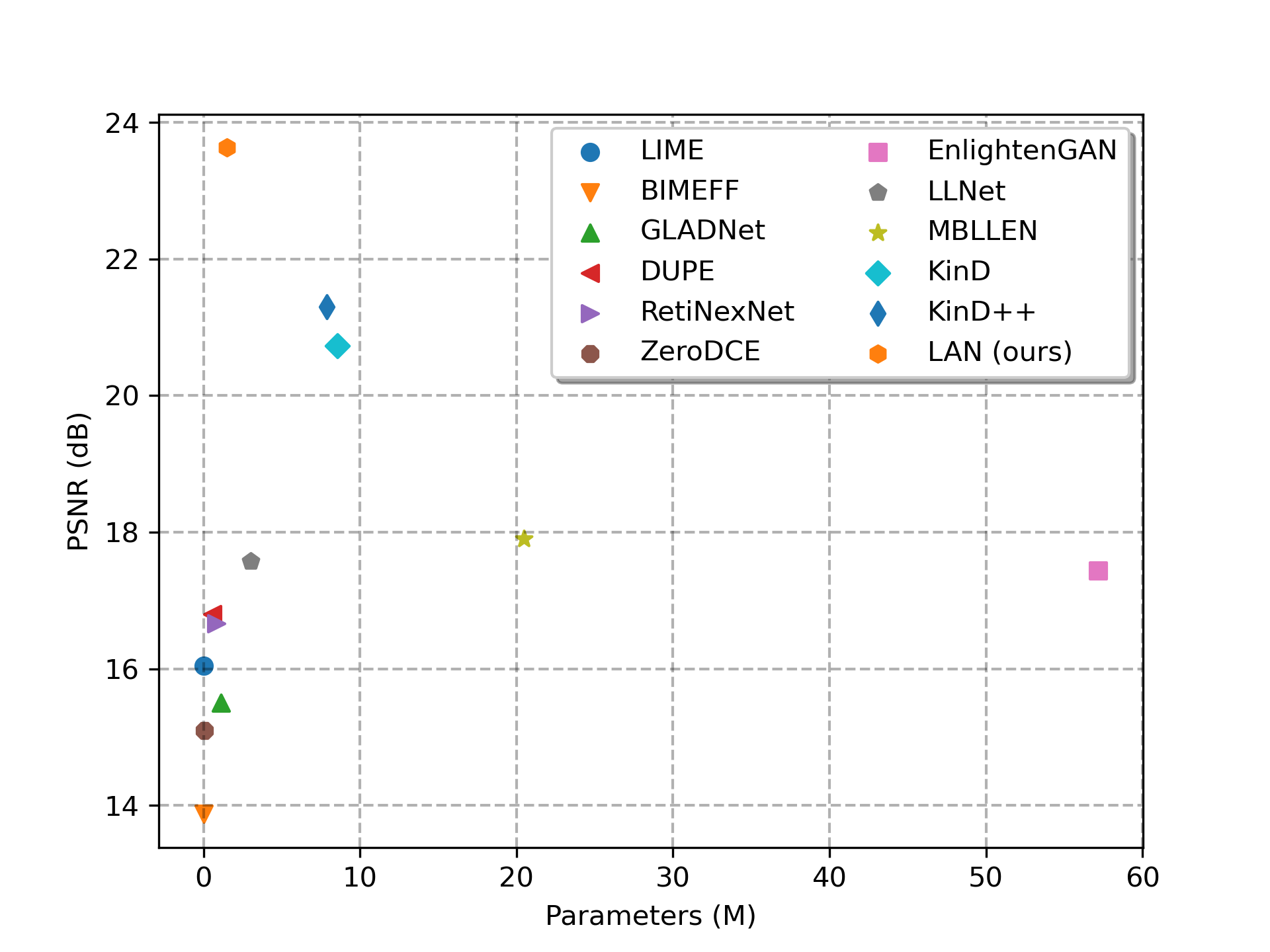}\\
\caption{The best PSNR-parameter trade-off of LAN (ours).
}
\label{fig1}
\end{figure}
In recent years, deep learning based methods have been dominant in the low-light image enhancement area and
convolution neural network (CNN) plays a fundamental role in this task~\cite{Lv2018MBLLEN,see_in_the_dark,RetinexNet,kind_kill_the_darkness}.
It enables learning powerful feature representations via massively stacked convolution layers.
To further boost the performance of convolution neural networks,
a lot of efforts have been done on exploring more robust architectures.
However, the inherent flaws of convolution neural networks (CNNs) have been always ignored.
Specifically, CNNs are convolution operation based on a local sliding window mechanism, which only focuses on the local relations of feature maps within the sliding window and makes it difficult to construct the global representation directly.
%然而，cnn存在的固有缺陷一直都被忽略。也就是，cnn是基于局部滑窗的卷积运算，这限制了它仅可关注滑窗内特征的局部关系，难以直接构建特征的全局表示。
%As show in Fig.~\ref{fig1} (a) and (b), the uneven color can be observed clearly due to lack of global information.

Recently, transformer-based architectures have been widely used in visual tasks, which can effectively aggregate feature maps global information.
However, visual transformer lacks the ability to constitute local feature representations, which is important for low-level visual tasks.
Recent works
~\cite{gao2021container,DBLP:journals/corr/abs-2105-03889}
attempt to alleviate this problem by coupling the local feature maps and global representations.
But the high computational cost of self-attention makes it difficult to cope with high resolution images directly even with the vast variants.

It is well known that attention mechanisms have been extensively used in previous literature, such as convolution block attention module (CBAM)~\cite{DBLP:conf/eccv/WooPLK18}, SimAM~\cite{Yang2021SimAMAS} and etc.
A natural idea is, can we design an attention mechanism with 3-D global weights to refine CNNs feature maps instead of directly aggregating whole feature maps global relationship to reduce the computational cost?
In light of this, this paper proposes a new attention mechanism named Linear Array Self-attention (LASA),
which can directly infer 3-D global attention weights from current feature maps and then in turn refine these feature maps.
%\textcolor[rgb]{1.00,0.00,0.00}{LASA is carefully designed as an attentional mechanism of auxiliary convolutional layer instead of directly constructing feature maps global representations.}
The refined feature maps can implicitly couple the local and global relationships by adjusting the local feature maps using global weights.
Specifically, LASA first encodes feature maps as two 2-D feature encodings along vertical and horizontal directions respectively.
Then, the global representations are constructed using the self-attention mechanism.
Finally, the 3-D global attention weights are generated by a multilayer perceptron (MLP) and a sigmoid activation function.
By this means, we can reduce the computational cost of self-attention from $O((H \times W)^2)$ to $O((H + W)^2)$.

As shown in Fig.~\ref{fig1}, this paper also develops a compact light-weight network named Linear Array Network (LAN) by adopting an autoencoder-like (AE) framework, which achieves the best trade-off between performance and the number of parameters compared with other methods.
Our LAN is constructed by stacking multiple Linear Array Block (LAB), where LAB is composed of LASA and convolutional layers.
Furthermore, multi-level features are also important for improving the performance of network.
Therefore, LAN uses skip-connection (SC) and global residual learning (GRL) to fuse shallow and deep features.
%In order to verify the effectiveness of the proposed method,
%we conduct extensive experiments in both RGB and RAW based low-light image enhancement tasks.
%LAN achieves superior performance to the existing SOTA methods with comparable amount of learnable parameters.
%To illustrate the rationality of each module of LAN, we perform a series of ablation experiments.
%At the same time, we conduct a comparative analysis with the existing 3-D weights attention mechanism and verify the effectiveness of LASA in different benchmark networks.
%At the end of the degradation experiment, we explore the application of comparative learning in low-light image enhancement tasks and analyze its advantages and limitations.
Our contributions can be summarized as:

(1) We propose a light-weight network for image restoration tasks and named Linear Array Network (LAN), which achieves SOTA results on both RGB and RAW based low-light image enhancement tasks and the best parameter-performance trade-off compared with other methods.

(2) Linear Array Self-attention (LASA) is proposed to enhance the global relationship construction ability of CNNs and handle the problem of the high computational complexity of self-attention.

(3) Extensive experiments prove that the validity of Linear Array Network (LAN) and Linear Array Self-attention (LASA).
At the end, a series of ablation experiments are constructed to validate the rationality of each module in LAN.

%(3) We explore the possibility of using contrastive learning in low-light image enhancement tasks and conduct a detailed analysis of its advantages and disadvantages through experiments.

\section{Related Work}

\subsection{Low-light Image Enhancement}
Modern low-light image enhancement tasks can be mainly split into two sub-directions: low-light RGB image enhancement and low-light RAW image reconstruction.
For low-light RGB image enhancement, LLNet~\cite{LLNet} first introduce an auto-encoder like CNN structure to handle this task.
Science then, various efforts have been done for using CNN-based structures~\cite{RetinexNet,Lv2018MBLLEN,zero_dce,kind_kill_the_darkness,enlightengan} to handle this task.
RetiNexNet~\cite{RetinexNet} incorporate the RetiNex theory into CNN to let the network deal with the reflection part and illumination part separately, then combine them together to get the final output.
Zero-DCE~\cite{zero_dce} propose a self-supervised strategy which only uses low-light image for network training.

Low-light RAW image reconstruction can be seen as a more challenging task, since the unprocessed low-light raw data often contains different type of in-camera noises and color aberrations.
%aberrations~\cite{camera_noise_model,CVPR_20_low_light_noise}.
Different from the traditional step-by-step methods
%Different from the traditional step-by-step methods~\cite{mobile_camera_imaging1,mobile_camera_imaging2}
for low-light RAW reconstruction, Chen \textit{et al.}~\cite{see_in_the_dark} first propose a U-Net
%~\cite{unet}
structure to handle this task, replacing the traditional manual-designed ISP pipeline, they also release the SID dataset which contains short-exposure low-light RAW images with their long-exposure RGB counterparts.
In order to reduce the number of parameters of the model, some works~\cite{LLPackNet,gu2019self} propose
%~\cite{LLPackNet,Real_Time_Dark_Restoration_Cvpr2021}
light-weight convolution neural networks (CNNs), however these methods reduce the quality of enhanced images at the same time.
Our LAN could achieve SOTA performance on both two tasks and better parameter-performance trade-off compared with previous methods.

\subsection{Attention Mechanism}
%Since ViT~\cite{ViT}, self-attention based model has been widely used in many computer vision tasks~\cite{DETR,swin,gao2021container}, also played a delighted effect on low-level vision tasks, like ~\cite{IPT_CVPR,TW} could handle multiple weather and image degradations with transformer model. Based on Swin-Transformer~\cite{swin}, Liang \textit{et al.} proposed Swin-IR~\cite{liang2021swinir} for single image super-resolution task. For image enhancement task, STAR~\cite{zhang2021star} adopt a long-short range module in transformer to handle white balance and photo retouching tasks.

%Compared to current attention based methods, such as Squeeze-and-Excitation (SE) attention~\cite{8701503}, Convolution Block Attention Module (CBAM)~\cite{DBLP:conf/eccv/WooPLK18} and so on.
%SE can only generate 1-D attention weight
%Although, CBAM can generate 3-D attention weight by exploit both spatial and channel-wise attention while it lack the ability of aggregation global information.
%LASA can augment the informative representations of feature maps by introduce 3-D global attention weights.
%Different from existing self-attention based aggregate global relationship methods or local-global coupling methods~\cite{gao2021container,DBLP:journals/corr/abs-2105-03889}.
%LASA have the ability of low computational complexity by compute the two 1-D feature encoding, and it can be easily plugged into existing convolution neural networks.

Attention mechanisms are effective methods to boost the performance
of CNNs via readjustment of feature maps, which enable CNNs to learn more reasonable
information.
%Humans nervous system likewise exist re-sculpt signal interneurons, such as inhibitory interneurons.
%Existing attention modules, calculated
%attention feature maps are 1-D or 2-D weights along either channel or spatial
%dimensions, and most of them lack global attention information, are still have
%some drawbacks.
Squeeze-and-Excitation (SE) attention learns the importance of different channels via learnable 1-D weights.
However, SE attention ignores the spatial importance on the feature maps.
To alleviate this problem, CBAM~\cite{DBLP:conf/eccv/WooPLK18}
%generates a 3-D weights in turn to refine feature maps
%emphasize meaningful features along those two principal dimensions
learns discriminative features
by exploiting both spatial and channel-wise attention.
In order to construct effective 3-D weights, SimAM~\cite{Yang2021SimAMAS} combines neuroscience
theories to propose a parameter-free attention module.
Self-attention~\cite{10.5555/3295222.3295349} mechanism is an effective method to aggregate global information.
However, it is proposed as an independent module which is hard to plug-and-play to CNNs, and the high computational cost makes it difficult to apply to high resolution images even with the vast variants.
Different from the self-attention based methods that calculate the global relationship of the feature maps directly, inspired by the attention mechanisms used in CNNs, our LASA constructs the global relationship of feature maps implicitly through 3-D weights.
To the best of our knowledge, LASA is the first work to introduce self-attention mechanism as an independent attention computing unit to low-light image enhancement tasks.

\begin{figure*}[htb]
\centering
    \begin{minipage}[c]{1\textwidth}
    \centering
    \includegraphics[width=7in]{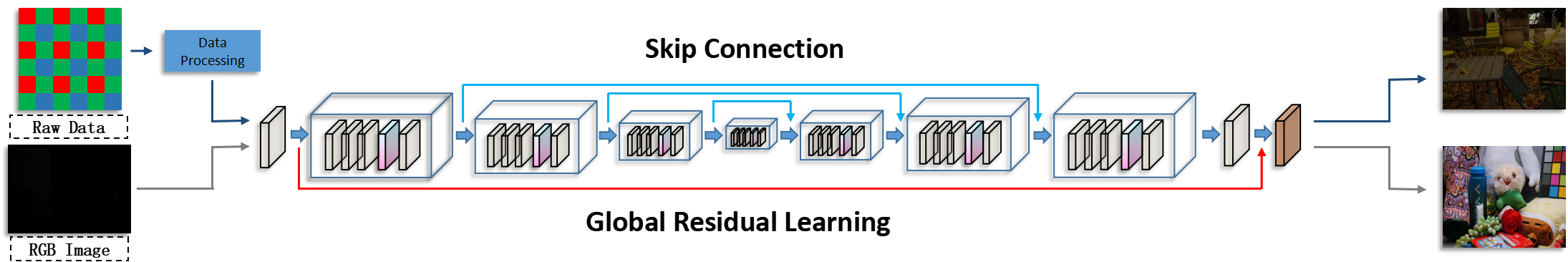}\\
    (a) Linear Array Network (LAN)
    \end{minipage}
%\rule[0.25\baselineskip]{\textwidth}{0.5pt}
\centering
    \begin{minipage}[c|c]{0.4\textwidth}
    \centering
    \includegraphics[width=2.8in]{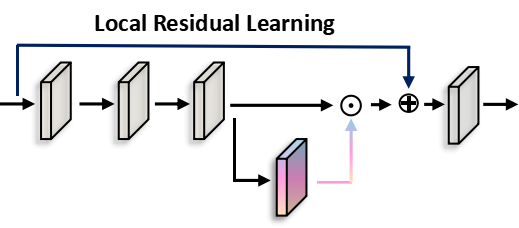}\\
    (b) Linear Array Block (LAB)
    \end{minipage}
    ~
    ~
    ~
    ~
    ~
    ~
    ~
    ~
\centering
    \begin{minipage}[c]{0.4\textwidth}
    \centering
    \includegraphics[width=3in]{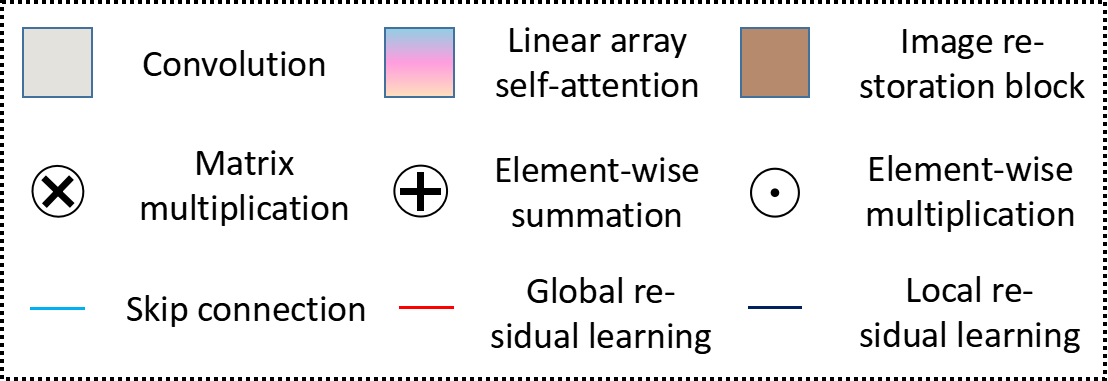}\\
    \end{minipage}

\centering
    \begin{minipage}[c]{1\textwidth}
    \centering
    \includegraphics[width=7in]{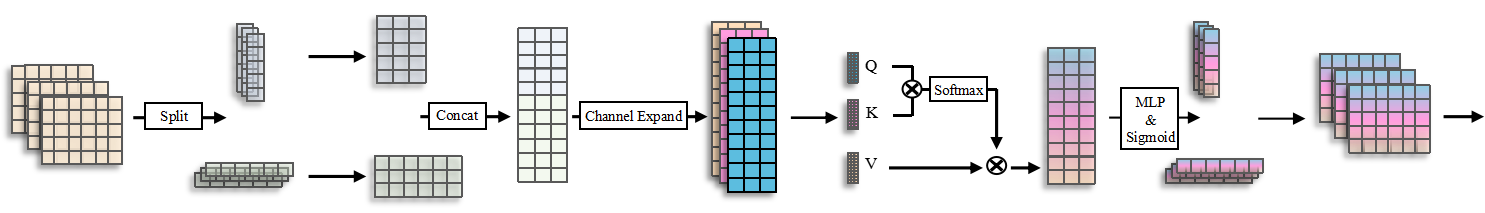}\\
    (c) Linear Array Self-attention (LASA)
    \end{minipage}

\caption{Network architecture of the proposed Linear Array Network (LAN). Zoom in for best view.
}
\label{fig2}
\end{figure*}

\section{Our Method}
%In this section, this paper first describe the low-light image enhancement pipeline.
%Then, we elaborate on the autoencoder-like (AE) low-light image enhancement network designed in this paper.
%Finally, the Linear array attention mechanism proposed in this article is introduced.

\subsection{Pipeline}
The current mainstream low-light image methods are mainly divided into two types, one is the method based on RGB images, and the other is the method of using RAW images to simulate image signal processor (ISP) .
RGB images usually only contain 8-bit information, while RAW images have 12-bit or higher bits information. Therefore, RAW images are more suitable for low-light image enhancement, especially for extremely low-light environments.
Since RAW image is unprocessed camera raw data, it needs to go through some data preprocessing operations before enhancement, such as black level correction and etc.
The RAW image enhancement pipeline is used in this paper same as Chen \textit{et al.}~\cite{see_in_the_dark}, except that the image enhancement network is redesigned.
For the enhancement pipeline of RGB images, this paper adopts an end-to-end approach like other low-light image enhancement methods based on RGB images.

\subsection{Linear Array Network}

To reduce the memory storage,
Linear Array Network (LAN) is an autoencoder-like (AE) network.
As show in Fig.~\ref{fig2}, LAN mainly contains three parts:
shallow feature extraction layers, linear array block (LAB), and finally the image restoration block.
Our shallow feature extraction layers consists of one regular convolution layer,
the extracted feature maps are used as input to the first LAB.
In order to obtain global dense features from the original input image, shallow features are also combined with deep features through global residual learning (GRL).
LAN also uses skip-connection (SC) to integrate the shallow (low-level features) and deep (high-level features) features with an addition operation.
Each linear array block consists of four regular convolution layers, one linear array self-attention (LASA), and local residual learning (LRL).
The linear array network uses seven linear array blocks, including three down-sampling modules and three up-sampling modules.
%Multi-level features~\cite{see_in_the_dark}  can effectively improve the performance of the network.
The image restore block is composed of a regular convolutional layer and a sub-pixel layer~\cite{7780576}.

\subsection{Linear Array Self-attention}

The detailed structure of the Linear Array Self-attention (LASA) is shown in the Fig.~\ref{fig2}.
It can be seen as an independent computing unit to enhance the expressive power of convolutional neural networks, and can be integrated into any other network as a plug-and-play module.
For a given feature map $\textbf{\emph{F}}~\in~\mathbb{R}~^{C \times H \times W}$,
LASA can directly infer a 3-D weights $\textbf{\emph{F}}_{attention}~\in~\mathbb{R}~^{C \times H \times W}$ with global information to refine the feature map.
The refined feature map can be computed as:
\begin{equation}
\textbf{\emph{F}}^{'} = \textbf{\emph{F}} \cdot \textbf{\emph{F}}_{attention},
\end{equation}
where $\cdot$ denotes element-wise multiplication,  $C,H,W$ are the number of channels, height and width of feature maps \textbf{\emph{F}} respectively.
As for the LASA,
we first encode feature map $\textbf{\emph{F}}~\in~\mathbb{R}~^{C \times H \times W}$
along the vertical and horizontal axes as a pair of 2-D feature encodings $\textbf{\emph{F}}_x~\in~\mathbb{R}~^{C \times 1 \times W }$
and
$\textbf{\emph{F}}_y~\in~\mathbb{R}~^{C  \times H \times 1}$,
which can be formulated as:

\begin{equation}
\textbf{\emph{F}}_{x}(i) = \frac{1}{H}\sum_{j=1}^H\textbf{\emph{F}}(i,j),
\end{equation}

\begin{equation}
\textbf{\emph{F}}_{y}(j) = \frac{1}{W}\sum_{i=1}^W\textbf{\emph{F}}(i,j).
\end{equation}

Next, we use matrix transformation operations to transform the sizes of feature map $\textbf{\emph{F}}_x~\in~\mathbb{R}~^{C \times W \times 1}$
and
$\textbf{\emph{F}}_y~\in~\mathbb{R}~^{C  \times 1 \times H}$
into
$\textbf{\emph{F}}_x~\in~\mathbb{R}~^{1 \times C \times W}$
and
$\textbf{\emph{F}}_y~\in~\mathbb{R}~^{1 \times C  \times H}$
respectively.
We concat feature map
$\textbf{\emph{F}}_x~\in~\mathbb{R}~^{1 \times C \times W}$ and $\textbf{\emph{F}}_y~\in~\mathbb{R}~^{1 \times C  \times H}$
along channel dimension and get a new feature map
$\textbf{\emph{F}}_{xy}~\in~\mathbb{R}~^{1 \times C \times (H+W)}$.
The number of channels of $\textbf{\emph{F}}_{xy}~\in~\mathbb{R}~^{1 \times C \times (H+W)}$ will be expanded to three times of the original
, and then divides it into three parts $\textbf{\emph{Q}}$, $\textbf{\emph{K}}$,
and $\textbf{\emph{V}}$
in the channel dimension.
Subsequently, the global relationship of the feature map is calculated, which can be formulated as:
\begin{equation}
\textbf{\emph{F}}_{global} = softmax(\textbf{\emph{Q}}\textbf{\emph{K}}^\mathrm{\emph{T}} ) \textbf{\emph{V}} + \textbf{\emph{F}}_{xy}.
\end{equation}
As shown above, after calculating the global relationship of the feature map, we adopt the residual learning strategy to facilitate the gradient flow.
At the end, the attention weights is computed as:
\begin{equation}
\textbf{\emph{F}}_{attention} = \sigma(MLP(\textbf{\emph{F}}_{global})).
\end{equation}
where MLP is a multi-layer perceptron, $\sigma$ is a sigmoid function.
\subsection{Loss Function}
Given an input image $\textbf{\emph{I}}_{in}$ and a ground turth image $\textbf{\emph{I}}_{gt}$, the loss function $L_{mix}$ for LAN consists of L1 loss $L^{\ell_1}$, MS-SSIM~\cite{1292216} loss $L_{MS-SSIM}$ and contrastive loss (CL)~\cite{wu2021contrastive} $L_{CL}$.
\begin{equation}
L^{\ell_1} = \frac{1}{N} \sum |\phi(\textbf{\emph{I}}_{in},w) - \textbf{\emph{I}}_{gt}|,
\end{equation}
where $N$ is the number of the pixels.
\begin{equation}
\resizebox{.91\linewidth}{!}{$
    \displaystyle
L^{MS-SSIM} = 1-\prod_{m-1}^{M}(\frac{2\mu_{p}\mu_{g}+c_1}{\mu^2_p+\mu^2_g+c_1})^{\alpha_m}(\frac{2\sigma_{pg}+c_2}{\sigma^2_p+\sigma^2_g+c_2})^{\beta_m},
$}
\end{equation}
%https://zhuanlan.zhihu.com/p/394785046
where $M$ represents images of different scales, $\mu_{p}$ and $\mu_{g}$ represent the mean values of the predicted image and the ground truth image, $\sigma_p$ and $\sigma_g$ represents the standard deviation of the predicted image and the ground truth image, and $\sigma_{pg}$ is the covariance between the two images. $\alpha_m$ and $\beta_m$ represent the weight coefficients between the two items, and $c_1$ and $c_2$ are two constants.
\begin{equation}
L^{CL}_{i} = \frac{D(G_i(\phi(\textbf{\emph{I}}_{in},w)),G_i(\textbf{\emph{I}}_{gt}))}{D(G_i(\phi(\textbf{\emph{I}}_{in},w)),G_i(\textbf{\emph{I}}_{in}))},
\end{equation}
where $D(x,y)$ is the $L_1$ distance, $L^{CL}_{i}$ is the $i$-th hidden features from the VGG model.
Therefore, the overall loss function $L^{MIX}$ used for LAN can be formulated as:
\begin{equation}
L^{MIX} = \lambda_1 \cdot L^{\ell_1} + \lambda_2 \cdot L^{MS-SSIM} + \lambda_3 \cdot \sum_i^{n} \cdot w_i  \cdot L^{CL}_{i}.
\label{Lmix}
\end{equation}
where $\lambda_1, \lambda_2 , \lambda_3, w_i$ are the weight coefficients used to make trade-off for the importance of the loss function $L^{MIX}$.
\section{Experiments and Results}

\begin{figure}[htb]
\centering
\begin{minipage}[c]{0.2\textwidth}
    \centering
        \begin{tikzpicture}[zoomboxarray, zoomboxes below, zoomboxarray inner gap=0.1cm, zoomboxarray columns=2, zoomboxarray rows=2]
            \node [image node] { \includegraphics[trim=0 0 0 0, clip, width=1.4in]{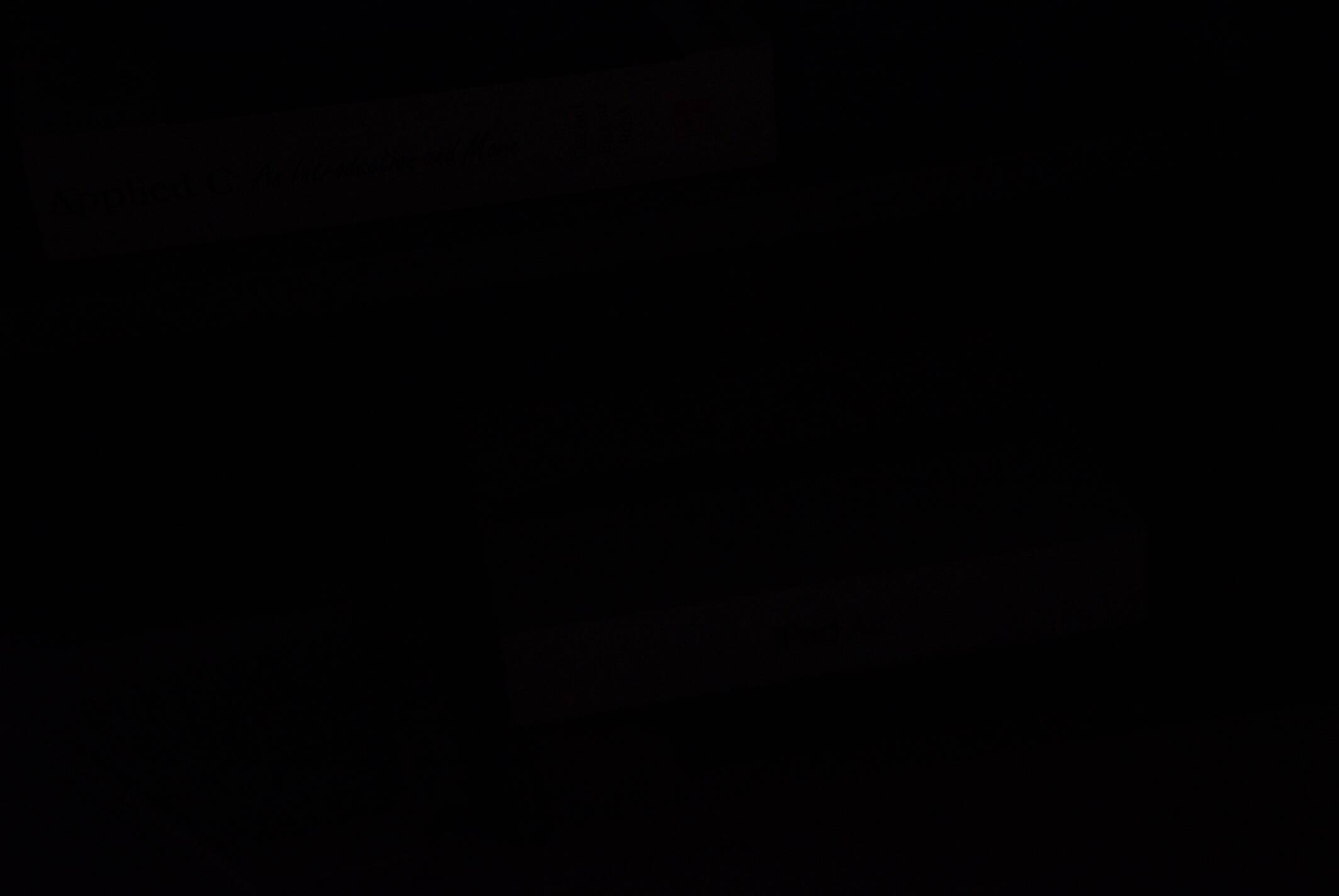} };
            \zoombox[magnification=4,color code=red]{0.10,0.08}
            \zoombox[magnification=4,color code=green]{0.45,0.1}
            \zoombox[magnification=4,color code=yellow]{0.60,0.3}
            \zoombox[magnification=4,color code=blue]{0.47,0.85}
        \end{tikzpicture}\\
            (a) Input
    \end{minipage}
\centering
\begin{minipage}[c]{0.2\textwidth}
    \centering
        \begin{tikzpicture}[zoomboxarray, zoomboxes below, zoomboxarray inner gap=0.1cm, zoomboxarray columns=2, zoomboxarray rows=2]
            \node [image node] { \includegraphics[trim=0 0 0 0, clip, width=1.4in]{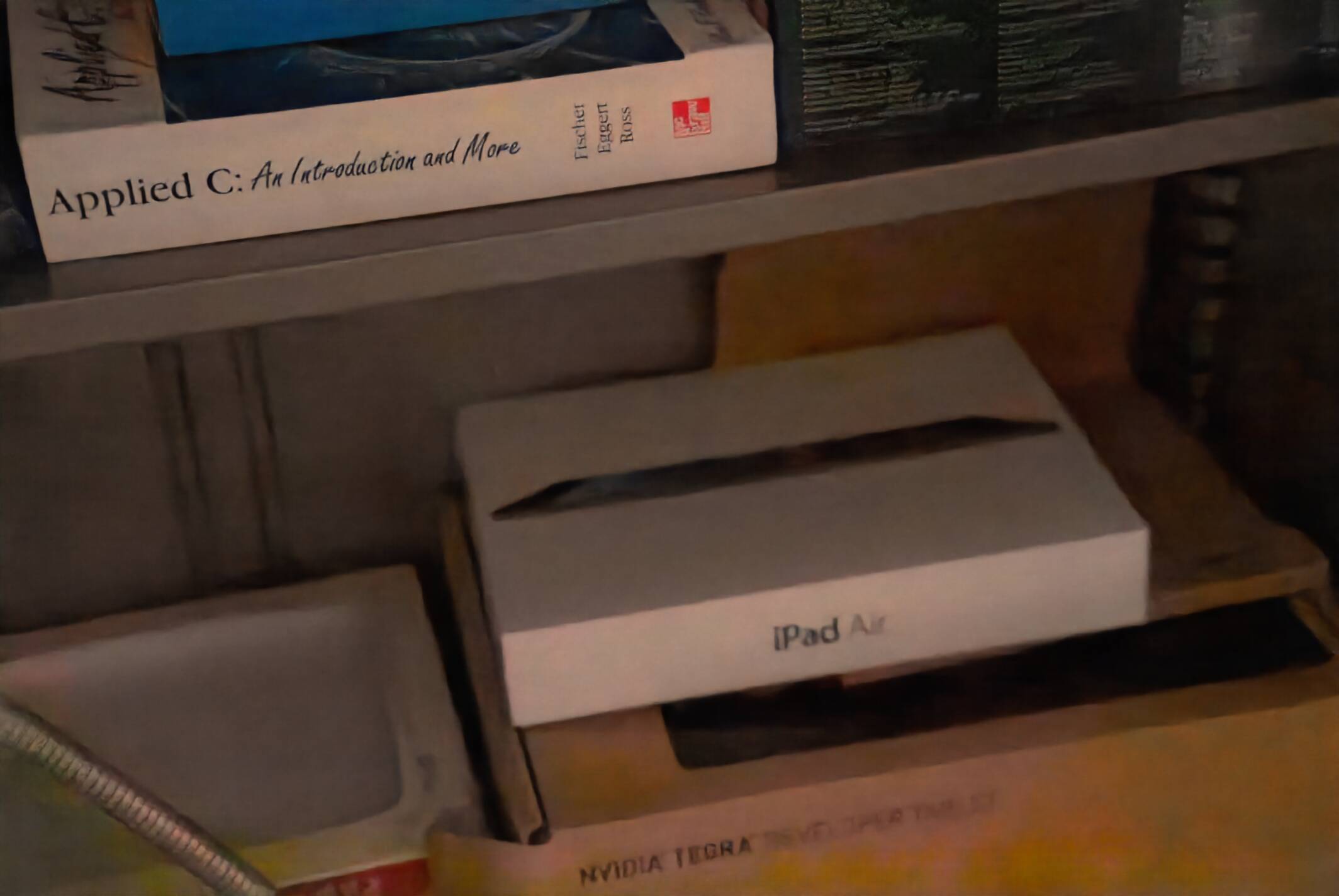} };
            \zoombox[magnification=4,color code=red]{0.10,0.08}
            \zoombox[magnification=4,color code=green]{0.45,0.1}
            \zoombox[magnification=4,color code=yellow]{0.60,0.3}
            \zoombox[magnification=4,color code=blue]{0.47,0.85}
        \end{tikzpicture}\\
            (b) LTS
\end{minipage}
\centering
\begin{minipage}[c]{0.2\textwidth}
    \centering
        \begin{tikzpicture}[zoomboxarray, zoomboxes below, zoomboxarray inner gap=0.1cm, zoomboxarray columns=2, zoomboxarray rows=2]
            \node [image node] { \includegraphics[trim=0 0 0 0, clip, width=1.4in]{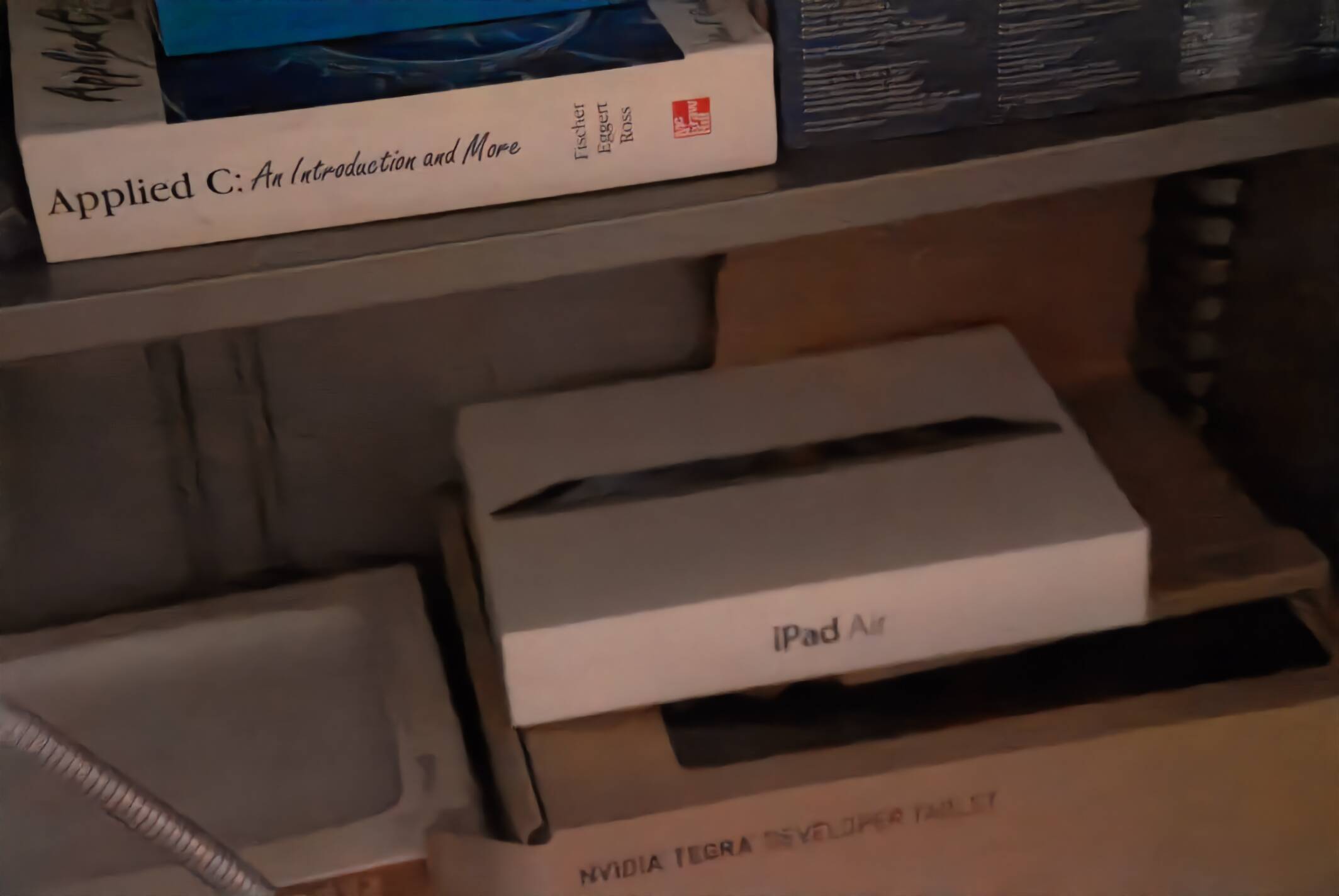} };
            \zoombox[magnification=4,color code=red]{0.10,0.08}
            \zoombox[magnification=4,color code=green]{0.45,0.1}
            \zoombox[magnification=4,color code=yellow]{0.60,0.3}
            \zoombox[magnification=4,color code=blue]{0.47,0.85}
        \end{tikzpicture}\\
            (c) LAN (Our)
\end{minipage}
\centering
\begin{minipage}[c]{0.2\textwidth}
    \centering
        \begin{tikzpicture}[zoomboxarray, zoomboxes below, zoomboxarray inner gap=0.1cm, zoomboxarray columns=2, zoomboxarray rows=2]
            \node [image node] { \includegraphics[trim=0 0 0 0, clip, width=1.4in]{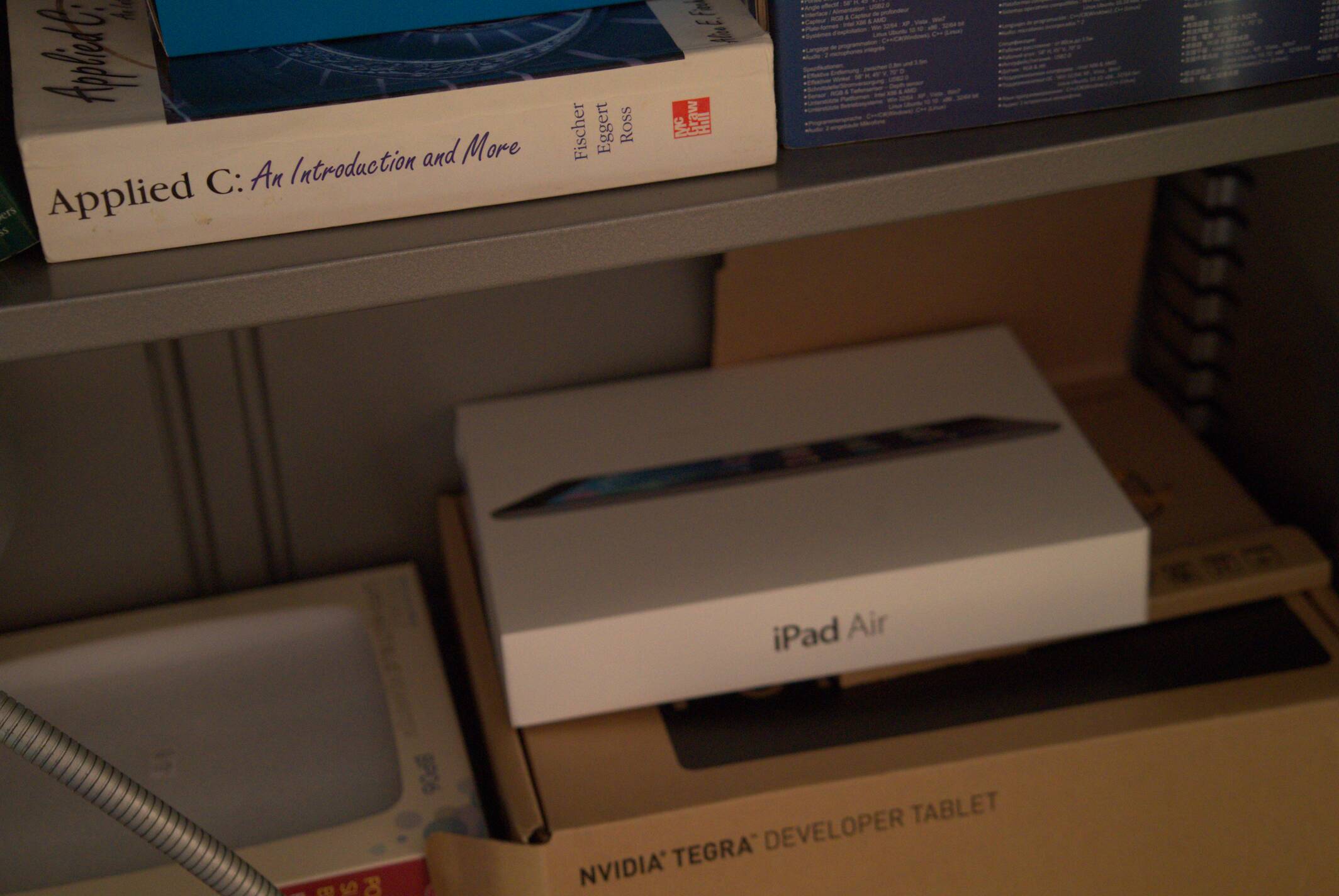} };
            \zoombox[magnification=4,color code=red]{0.10,0.08}
            \zoombox[magnification=4,color code=green]{0.45,0.1}
            \zoombox[magnification=4,color code=yellow]{0.60,0.3}
            \zoombox[magnification=4,color code=blue]{0.47,0.85}
        \end{tikzpicture}\\
            (d) GT
\end{minipage}
\caption{Visual results comparison on SID dataset. Zoom in for best view.
}
\label{fig3}
\end{figure}

\begin{table*}[htb]
\label{tabat}
\footnotesize
\tabcolsep 11pt
\begin{tabular*}{\textwidth}{cccccccc}
\toprule
  \multicolumn{1}{c}{LOL} &  \multicolumn{1}{c}{PSNR~(dB)~$\uparrow$}  & \multicolumn{1}{c}{SSIM~$\uparrow$} & \multicolumn{1}{c}{GMSD~$\downarrow$} & \multicolumn{1}{c}{NLPD~$\downarrow$} & \multicolumn{1}{c}{NIQE~$\downarrow$} &  \multicolumn{1}{c}{DISTS~$\downarrow$} &  \multicolumn{1}{c}{Param~$\downarrow$} \\\hline
    \multicolumn{1}{c}{LIME} & \multicolumn{1}{c}{16.0458} & \multicolumn{1}{c}{0.4834} & \multicolumn{1}{c}{0.1541} & \multicolumn{1}{c}{0.5129} &  \multicolumn{1}{c}{10.8926} &  \multicolumn{1}{c}{0.1806} &  \multicolumn{1}{c}{-} \\

    \multicolumn{1}{c}{BIMEF}   & 13.8752 &  0.5936 & 0.0953 & 0.3645 &  9.8083 &  0.1878 & -  \\

    \multicolumn{1}{c}{GLADNet} & 15.5045 & 0.6247 & 0.2035 & 0.5169 &  16.9071 &  0.3072 & 1.128 M \\

    \multicolumn{1}{c}{DUPE} &  16.7975 & 0.5187 & 0.1675 & 0.5936 &  10.4406 & 0.1794 & 0.5652 M \\

    \multicolumn{1}{c}{RetiNexNet}& 16.6691 & 0.4909 & 0.1549 & 0.5799 &  8.9796 & 0.2450 & 0.838 M \\

    \multicolumn{1}{c}{ZeroDCE}& 15.0924 & 0.5093 & 0.1646 & 0.4878 & 10.4976 &  0.1891 & \textbf{0.0794 M} \\

    \multicolumn{1}{c}{EnlightenGAN}& 17.4412 & 0.6744 & 0.1046 & 0.3674 &  14.8651 &  0.1638 & 57.17 M \\

    \multicolumn{1}{c}{LLNet}&   17.5777 & 0.6819 & 0.1485 & 0.4837 & 14.2112 & 0.2266 & 3.012 M \\

    \multicolumn{1}{c}{MBLLEN}&  17.9006 & 0.7020 & 0.1160 & 0.3447 & 14.7112 & 0.1448 & 20.4746 M \\

    \multicolumn{1}{c}{KinD}&  20.7261 & 0.8103 & 0.0888 & 0.3187 & 10.7841 & 0.1126 & 8.540 M  \\

    \multicolumn{1}{c}{KinD++}&  21.3003 & 0.8226 & 0.0960 & 0.3174 & 11.3194 & 0.1169 & 7.8912 M \\

    \multicolumn{1}{c}{LAN (ours)}& \textbf{23.6324} & \textbf{0.8444} & \textbf{0.0670} & \textbf{0.2683} & \textbf{9.9427} & \textbf{0.0801} & 1.49 M  \\
    \bottomrule

\end{tabular*}
\caption{Quantitative evaluation of low-light image enhancement algorithms on LOL dataset.The best results are highlighted in bold.}
\label{tab1}
\end{table*}

\subsection{Experimental Setup}

\textbf{Dataset.}
Our evaluation experiments are constructed on two widely used low-light datasets: LOL dataset (RGB Images) and SID dataset (RAW Images). The LOL dataset contains 485 training samples and 15 testing samples. SID datasets consist of two sub-datasets: SID-Sony (SIDS) and SID-Fuji (SIDF). SIDS contains 231 high-exposure images and 2697 short-exposure images, and SIDF contains 193 high-exposure images and 2397 short-exposure images. Each high-exposure image corresponds to different short-exposure images in SIDS and SIDF.
%The difference between SID dataset and LOL dataset is that SID is a RAW format image that is captured by camera sensor while LOL is RGB format image.

\textbf{Metrics.}
To evaluate the performance of our method, we have adopted the classical image quality assessment methods: Peak Signal to Noise Ratio (PSNR) and the Structural Similarity index (SSIM) as the evaluation metrics. They are usually used as the criteria to evaluate image quality in image restoration tasks. In order to conduct quantitative analysis objectively, we have also adopted other evaluation metrics, including
PSNR, SSIM
%~\cite{wang2004image}
, gradient magnitude similarity deviation (GMSD)~\cite{xue2013gradient}, normalized Laplacian pyramid (NLPD)~\cite{laparra2016perceptual}, natural image quality
evaluator (NIQE)~\cite{mittal2012making}, deep image structure and texture
similarity (DISTS)~\cite{ding2020image}.

\textbf{Implementation Details.}
Our LAN is implemented by pytorch with one NVIDIA GeForce RTX 2080Ti GPU.
The model is trained on Adam optimizer with default parameters.
The epoch is set to 4000 and 2000 for SID and LOL, respectively.
For these two datasets, we set the initial learning rate to 0.0001, and reduce the learning rate to one-tenth of the original in half of the training epoch.
The batch size is set to 1 for both LOL and SID datasets.
The data augmentation methods used during training are random rotation and flipping.

\subsection{Quantitative Evaluation}

We evaluate our method on two widely-adopted datasets, LOL  and SID respectively.
The compared methods on LOL dataset including LIME~\cite{7782813}, BIMEF~\cite{ying2017bio}, GLADNet~\cite{LLenhance_glad_net}, DUPE~\cite{wang2019underexposed}, RetiNexNet~\cite{RetinexNet},
ZeroDCE~\cite{zero_dce}, EnlightenGAN~\cite{enlightengan}, LLNet~\cite{LLNet}, MBLLEN~\cite{Lv2018MBLLEN}, KinD~\cite{kind_kill_the_darkness} and KinD++~\cite{zhang2021beyond}.
In addition, to evaluate the
enhancement effectiveness of our proposed method,
the models are also evaluated on extremely low-light dataset (SID).
We compare our models with several state-of-the-art
methods: LTS~\cite{see_in_the_dark}, DID~\cite{maharjan2019improving}, SGN~\cite{gu2019self}, LLPAckNet~\cite{LLPackNet}, EEMEFN~\cite{zhu2020eemefn}, LDC~\cite{LDC_CVPR2020}.

Tab.~\ref{tab1}, Tab.~\ref{tab2} and Tab.~\ref{tab3} respectively reports quantitative results on LOL, SIDS and SIDF datasets.
It can be seen that LAN achieves a better trade-off between model parameters and performance compared with other methods.
From Tab.~\ref{tab1}, we can find that LAN outperforms all the other competitors by a large margin under multiple evaluation metrics.
As shown in Tab.~\ref{tab2} and Tab.~\ref{tab3}, we can see that LAN also has great advantages compared to other methods in the RAW image based low-light image enhancement task.
Although LAN has a slightly lower SSIM index than LDC by 0.06, the number of parameters of LDC is 5 times higher than that of LAN.

%Let us compare the model with the existing methods, and
%the results are shown in Table 1.It reports the numerical results among the competitors on the LOL dataset. For each testing low-light image,
%there is a “normal”-light correspondence. Thus, the correspondence can be taken as reference to measure PSNR and
%SSIM.From the numbers, we see our model
%significantly outperform all the other methods. In terms of
%the non-reference metric NIQE and DeltaE, our model also
%show their superiority over the others by alarge margin.
%the proposed
%model approach achieves the best performance with an average PSNR score of 23.5710 dB and SSIM score of 0.8444,
%which exceed the second-best approach (LLNet) by 3.993 dB
%(23.5710-17.953) on PSNR and 0.103 (0.807-0.704) on SSIM.
%
%From the definition of LOE
%(Wang et al.2013), we can see that the reference is cru-
%cial to quantitatively measuring the quality of enhancement.
%However, it is problematic to calculate LOE using a low-light  image  as the input itself when no enhancement is performed.To more appro-
%priately reflect the enhancement quality in terms of LOE,
%a suitable reference matters.We use correspondence images for reference,meanwhile,In terms of non reference measures NIQE and DELTA, our model also have great advantages.

\begin{table}[htb]
\centering
\begin{tabular}{llll}
\hline
\multicolumn{1}{c}{SIDS} &  \multicolumn{1}{c}{PSNR~(dB)~$\uparrow$}  & \multicolumn{1}{c}{SSIM~$\uparrow$} &  \multicolumn{1}{c}{Param~$\downarrow$} \\\hline
    %\multicolumn{1}{c}{CAN} & \multicolumn{1}{c}{27.40} & \multicolumn{1}{c}{0.792} & \multicolumn{1}{c}{-} \\

    \multicolumn{1}{c}{LTS}&  \multicolumn{1}{c}{28.88} & \multicolumn{1}{c}{0.787} & \multicolumn{1}{c}{7.7 M}  \\

    \multicolumn{1}{c}{DID}& \multicolumn{1}{c}{28.41} & \multicolumn{1}{c}{0.780} & \multicolumn{1}{c}{2.5 M}  \\

    \multicolumn{1}{c}{SGN}&  \multicolumn{1}{c}{28.91} & \multicolumn{1}{c}{0.789} & \multicolumn{1}{c}{3.5 M } \\

    \multicolumn{1}{c}{LLPAckNet}&  \multicolumn{1}{c}{27.83} & \multicolumn{1}{c}{0.75} & \multicolumn{1}{c}{\textbf{1.16 M}}  \\
%    \multicolumn{1}{c}{MEF} & \multicolumn{1}{c}{29.43} & \multicolumn{1}{c}{0.791} & \multicolumn{1}{c}{-} \\
    %\multicolumn{1}{c}{DCE}&  \multicolumn{1}{c}{26.53} & \multicolumn{1}{c}{0.73} & \multicolumn{1}{c}{0.79 M}  \\
    \multicolumn{1}{c}{EEMEFN} &  \multicolumn{1}{c}{29.60} & \multicolumn{1}{c}{0.795} & \multicolumn{1}{c}{40.713M}\\
    \multicolumn{1}{c}{LDC}&  \multicolumn{1}{c}{29.56} & \multicolumn{1}{c}{\textbf{0.799}} & \multicolumn{1}{c}{8.6 M}  \\

    \multicolumn{1}{c}{LAN (ours)}&  \multicolumn{1}{c}{\textbf{30.17}} & \multicolumn{1}{c}{0.793} & \multicolumn{1}{c}{1.48 M}  \\\hline
\end{tabular}
\caption{Quantitative evaluation of low-light image enhancement algorithms on SIDS dataset.The best results are highlighted in bold.}
\label{tab2}
\end{table}

\begin{table}[htb]
\centering
\begin{tabular}{llll}
\hline
\multicolumn{1}{c}{SIDF} &  \multicolumn{1}{c}{PSNR~(dB)~$\uparrow$}  & \multicolumn{1}{c}{SSIM~$\uparrow$} &  \multicolumn{1}{c}{Param~$\downarrow$} \\\hline
    %\multicolumn{1}{c}{CAN} & \multicolumn{1}{c}{25.71} & \multicolumn{1}{c}{0.719} & \multicolumn{1}{c}{-} \\
    \multicolumn{1}{c}{LTS}&  \multicolumn{1}{c}{26.61} & \multicolumn{1}{c}{0.680} & \multicolumn{1}{c}{7.7 M}  \\
    \multicolumn{1}{c}{SCG} & \multicolumn{1}{c}{26.90} & \multicolumn{1}{c}{0.683} & \multicolumn{1}{c}{3.5 M} \\
    \multicolumn{1}{c}{LLPackNet} & \multicolumn{1}{c}{24.13} & \multicolumn{1}{c}{0.59} & \multicolumn{1}{c}{\textbf{1.16 M}} \\
    \multicolumn{1}{c}{EEMEFN} &  \multicolumn{1}{c}{27.38}& \multicolumn{1}{c}{0.723} & \multicolumn{1}{c}{40.713M}\\
%    \multicolumn{1}{c}{MEF}&  \multicolumn{1}{c}{27.21} & \multicolumn{1}{c}{0.719} & \multicolumn{1}{c}{-}  \\
    \multicolumn{1}{c}{LDC} & \multicolumn{1}{c}{26.70} & \multicolumn{1}{c}{0.681} & \multicolumn{1}{c}{8.6 M } \\
    \multicolumn{1}{c}{LAN (ours)}&  \multicolumn{1}{c}{\textbf{28.02}} & \multicolumn{1}{c}{\textbf{0.720}} & \multicolumn{1}{c}{1.49 M}  \\\hline
\end{tabular}
\caption{Quantitative evaluation of low-light image enhancement algorithms on SIDF dataset.The best results are highlighted in bold.}
\label{tab3}
\end{table}

\begin{figure*}[!t]
\centering
    \begin{minipage}[c]{0.17\textwidth}
    \centering
    \includegraphics[width=1.2in]{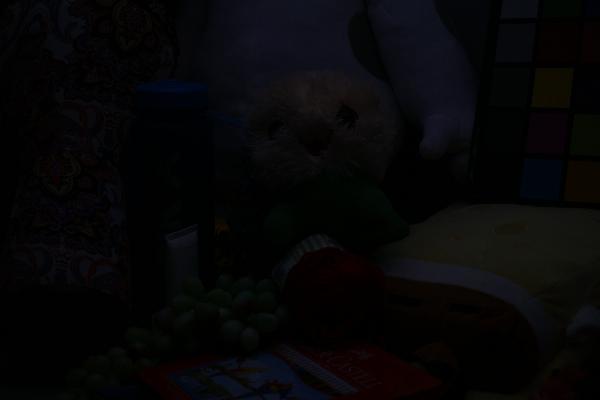}\\
    (a) Input
    \end{minipage}
\centering
    \begin{minipage}[c]{0.17\textwidth}
    \centering
    \includegraphics[width=1.2in]{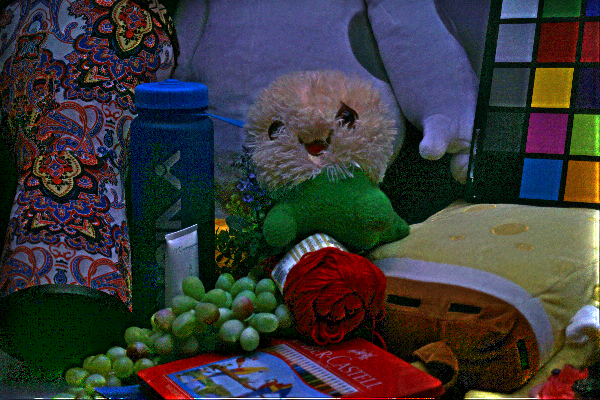}\\
    (b) LIME
    \end{minipage}
\centering
    \begin{minipage}[c]{0.17\textwidth}
    \centering
    \includegraphics[width=1.2in]{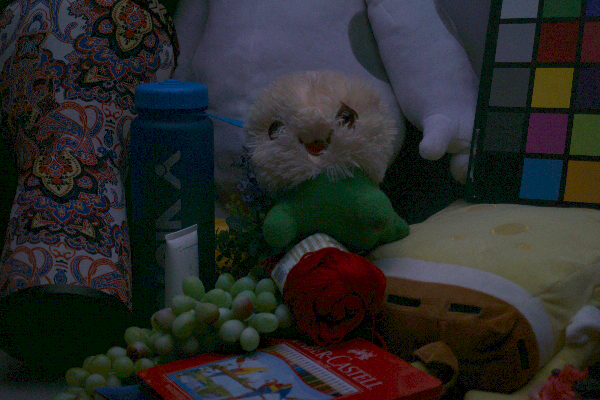}\\
    (c) BIMEF
    \end{minipage}
\centering
    \begin{minipage}[c]{0.17\textwidth}
    \centering
    \includegraphics[width=1.2in]{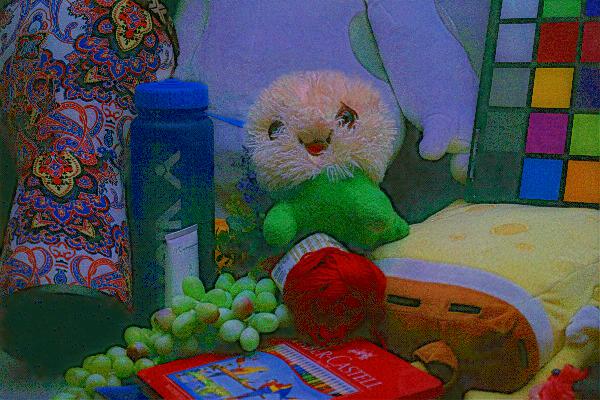}\\
    (d) RetinexNet
    \end{minipage}
\centering
    \begin{minipage}[c]{0.17\textwidth}
    \centering
    \includegraphics[width=1.2in]{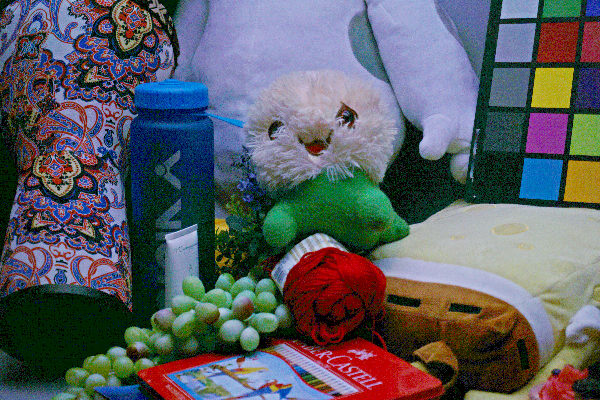}\\
    (e) ZeroDCE
    \end{minipage}
\centering
    \begin{minipage}[c]{0.17\textwidth}
    \centering
    \includegraphics[width=1.2in]{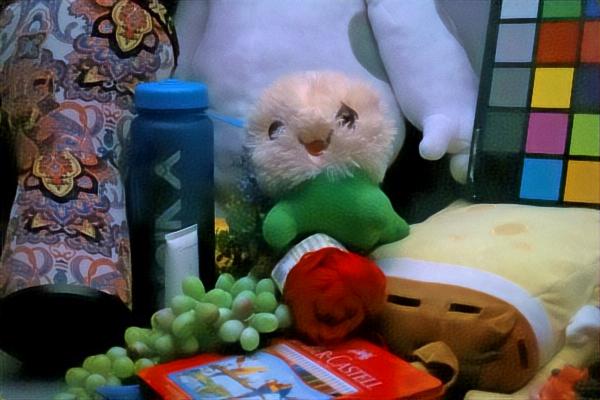}\\
    (f) MBLLEN
    \end{minipage}
\centering
    \begin{minipage}[c]{0.17\textwidth}
    \centering
    \includegraphics[width=1.2in]{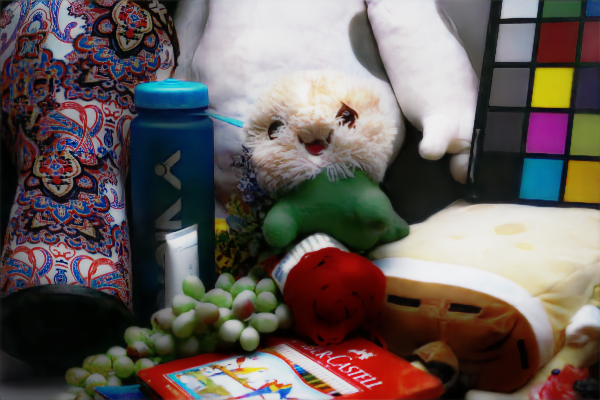}\\
    (g) KinD
    \end{minipage}
\centering
    \begin{minipage}[c]{0.17\textwidth}
    \centering
    \includegraphics[width=1.2in]{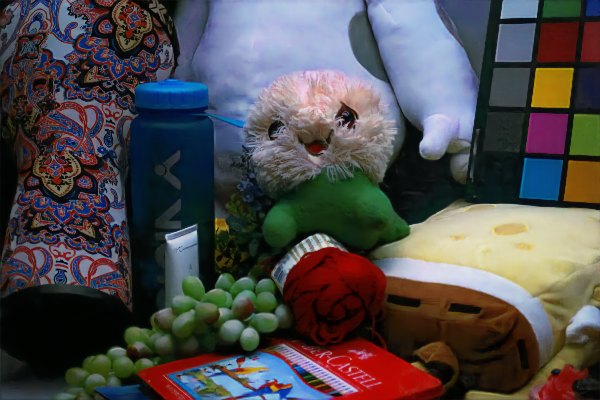}\\
    (h) KinD++
    \end{minipage}
\centering
    \begin{minipage}[c]{0.17\textwidth}
    \centering
    \includegraphics[width=1.2in]{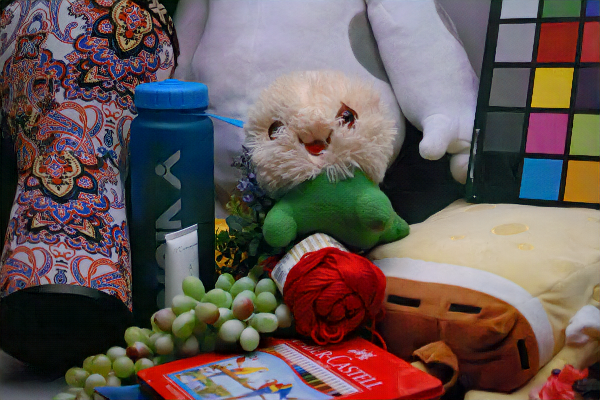}\\
    (i) LAN (Ours)
    \end{minipage}
\centering
    \begin{minipage}[c]{0.17\textwidth}
    \centering
    \includegraphics[width=1.2in]{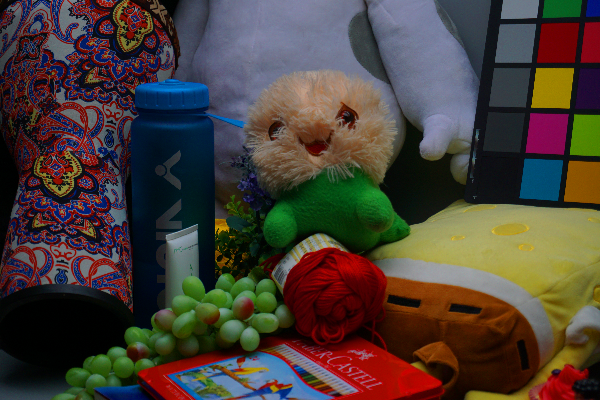}\\
    (j) GT
    \end{minipage}

\caption{Visual results comparison on LOL dataset. Zoom in for best view.
}
\label{fig4}
\end{figure*}

\begin{table}[htb]
\centering
\begin{tabular}{ll|lll}
\hline
\multicolumn{1}{c}{Model}  &  \multicolumn{1}{c|}{LASA}  &  \multicolumn{1}{c}{PSNR~(dB) $\uparrow$} & \multicolumn{1}{c}{SSIM $\uparrow$}  \\\hline
\multicolumn{1}{c}{Base}& \multicolumn{1}{c|}{-} & \multicolumn{1}{c}{15.485} & \multicolumn{1}{c}{0.569}     \\
\multicolumn{1}{c}{Base+SC}&\multicolumn{1}{c|}{-}& \multicolumn{1}{c}{28.702} & \multicolumn{1}{c}{0.784}   \\
\multicolumn{1}{c}{Base+SC+GRL}&\multicolumn{1}{c|}{-}& \multicolumn{1}{c}{28.907}&\multicolumn{1}{c}{0.785}   \\
\multicolumn{1}{c}{Base+SC+GRL+LRL}&\multicolumn{1}{c|}{-}& \multicolumn{1}{c}{28.673} & \multicolumn{1}{c}{0.784}     \\\hline
\multicolumn{1}{c}{Base+SC+GRL}& \multicolumn{1}{c|}{$\surd$} & \multicolumn{1}{c}{30.075} & \multicolumn{1}{c}{\textbf{0.794}}  \\
\multicolumn{1}{c}{Base+SC+GRL+LRL}& \multicolumn{1}{c|}{ $\surd$} &\multicolumn{1}{c}{\textbf{30.170}}&\multicolumn{1}{c}{0.793} \\\hline
\end{tabular}
\caption{Ablation study on the design of Linear Array Network.}
\label{ab1}
\end{table}

\begin{table}[htb]
\centering
\begin{tabular}{ll|ll}
\hline
\multicolumn{1}{c}{Model} & \multicolumn{1}{c|}{Attention}   &  \multicolumn{1}{c}{PSNR~(dB) $\uparrow$}  & \multicolumn{1}{c}{SSIM $\uparrow$}   \\\hline
\multicolumn{1}{c}{LAN} &  \multicolumn{1}{c|}{-}  & \multicolumn{1}{c}{28.673} & \multicolumn{1}{c}{0.784}     \\
%LAAN & SE  & 29.791 (+) & 0.787 (+)     \\
\multicolumn{1}{c}{LAN} & \multicolumn{1}{c|}{CBAM}  & \multicolumn{1}{c}{29.409 (0.736$\uparrow$)} & \multicolumn{1}{c}{0.784}      \\
\multicolumn{1}{c}{LAN} & \multicolumn{1}{c|}{SimAM}  & \multicolumn{1}{c}{29.008 (0.335$\uparrow$)} & \multicolumn{1}{c}{0.785 (0.001$\uparrow$)}      \\
\multicolumn{1}{c}{LAN} & \multicolumn{1}{c|}{LASA}  & \multicolumn{1}{c}{\textbf{30.170} (1.497$\uparrow$)} & \multicolumn{1}{c}{\textbf{0.793} (0.009$\uparrow$)}    \\\hline
\multicolumn{1}{c}{LTS} & \multicolumn{1}{c|}{-}  & \multicolumn{1}{c}{28.88} & \multicolumn{1}{c}{0.787}     \\
%LTS & SE   & 15.485 & 0.569     \\
\multicolumn{1}{c}{LTS} & \multicolumn{1}{c|}{CBAM}   & \multicolumn{1}{c}{27.334 (1.546$\downarrow$)} & \multicolumn{1}{c}{0.765 (0.022$\downarrow$)}     \\
\multicolumn{1}{c}{LTS} & \multicolumn{1}{c|}{SimAM}   & \multicolumn{1}{c}{29.127 (0.247$\uparrow$)} & \multicolumn{1}{c}{0.785 (0.002$\downarrow$)}     \\
\multicolumn{1}{c}{LTS} & \multicolumn{1}{c|}{LASA}   & \multicolumn{1}{c}{\textbf{30.035} (1.155$\uparrow$)} & \multicolumn{1}{c}{\textbf{0.792} (0.005$\uparrow$)}     \\\hline
\end{tabular}
\caption{Comparisons of different attention mechanism.}
\label{ab2}
\end{table}

\subsection{Qualitative Evaluation}

We compare our approach with other methods on both SIDS and LOL datasets and provide the qualitative results in Fig.~\ref{fig3} and Fig.~\ref{fig4}.
As shown in Fig.~\ref{fig3}, it is clear that our method have sharper details and more natural color constancy.
Especially in the carton part of the picture, due to the ability to build global relationships, the color consistency is higher and no visual artifacts appear.
From Fig.~\ref{fig4}, we can see that compared with other methods, the enhanced image of LAN has lower noise, higher contrast, clearer texture details, and more realistic image content.
\begin{figure}[htb]
\centering
    \begin{minipage}[c]{0.2\textwidth}
    \centering
    \includegraphics[width=1.3in]{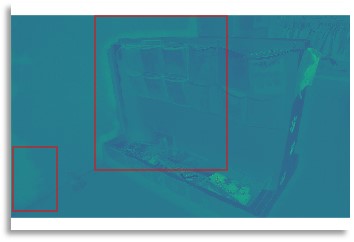}\\
    %(a) Low-light Image
    \end{minipage}
\centering
    \begin{minipage}[c]{0.2\textwidth}
    \centering
    \includegraphics[width=1.3in]{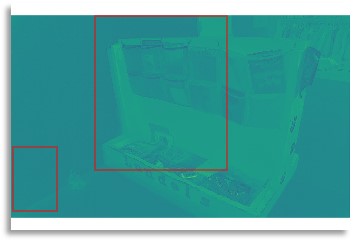}\\

    \end{minipage}
%\hspace*{0.01cm}
    \begin{minipage}[c]{0.2\textwidth}
    \centering
    \includegraphics[width=1.3in]{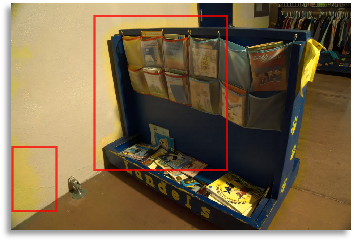}\\
    (a) CNNs
    \end{minipage}
\centering
    \begin{minipage}[c]{0.2\textwidth}
    \centering
    \includegraphics[width=1.3in]{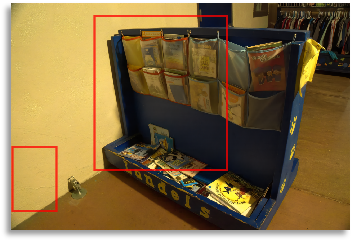}\\
    (b) CNNs+LASA
    \end{minipage}

\caption{Visual comparison of with and without LASA.The upper part of the figure are the feature maps visualization results, and the lower part are the enhanced image results.}
\label{fig5}
\end{figure}
\subsection{Ablation Experiment}
In order to further verify the effectiveness of different elements on the network including skip-connection (SC), global residual learning (GRL), local residual learning (LRL) and LASA proposed in this paper, we constructed an ablation study. As shown in Tab.~\ref{ab1}, the base network is the baseline of our ablation experiment, which mainly consists of three downsampling linear array blocks, three upsampling linear array blocks, one linear array block, shallow feature extraction layers and one image restoration block. In the LASA column of Tab.~\ref{ab1}, '-' means LASA is not used, and '$\surd$' means LASA is used.

Subsequently, we gradually add different modules to the base network:
(1) \textbf{base+SC}: Add the skip-connection (SC) operation into the baseline network.
(2) \textbf{base+SC+GRL}: Add the skip-connection (SC) and the global residual learning (GRL) operation into the baseline network.
(3) \textbf{base+SC+GRL+LRL}: Add the skip-connection (SC), gloabl residual learning (GRL) and local residual learning (LRL) operation into the baseline network.
(4) \textbf{base+SC+GRL+LASA}: Add the skip-connection (SC), gloabl residual learning (GRL) and linear array self-attention into the baseline network.
(5) \textbf{base+SC+GRL+LRL+LASA}: Add the skip-connection (SC), gloabl residual learning (GRL), local residual learning (LRL) and linear array self-attention into the baseline network.
As we can see from Tab.~\ref{ab1}, when we gradually add different modules to the baseline network, the performance of the model will gradually improve. Especially after using LASA, the PSNR will increase by more than 1 dB. Therefore, the effectiveness of the LASA can be fully proved.

Furthermore, to verify the effectiveness of LASA, we compared it with the commonly used 3-D wights attention mechanisms CBAM~\cite{DBLP:conf/eccv/WooPLK18} and SimAM~\cite{Yang2021SimAMAS}.
%\textcolor[rgb]{1.00,0.00,0.00}{
%%The method used here is the same as the other method used in this article that applies the 3-D attention weights to refine CNN feature maps.
%As shown in Tab.~\ref{ab2}, we retrain all of the methods.
%For the LAN+Attention experiments, we directly use CBAM and SimAM to replace the LASA of the original network.
%For the LTS based methods, we add an attention mechanism to the last layer of each module.}
As shown in the upper part of Tab.~\ref{ab2}, using CBAM and SimAM in the LAN will bring a PSNR increase of 0.736 dB and 0.335 dB respectively, while using LASA will bring a PSNR increase of 1.497 dB.
We also use the LTS~\cite{see_in_the_dark} as the baseline method for experimental comparison.
From the bottom half of Tab.~\ref{ab2}, we can see that the PSNR of LTS decreased by 1.546 dB after using CBAM.
SimAM can increase the PSNR slightly by 0.247 dB and the PSNR goes up significantly by 1.155 dB when using LASA.
Based on the experimental analysis above, LASA can be plugged into any network structure and therefore improves performance of network.

We also compare the performance of CNNs+LASA against using
only CNNs, with LAN as the baseline network. As can be
seen from Fig.~\ref{fig5}, the method using only CNNs has
a serious color imbalance problem, which can be effectively
alleviated by using LASA.

\section{Conclusion}
In this paper, we propose Linear Array Network (LAN) for low-light image enhancement, which consists autoencoder-like (AE) network and Linear Array Self-attention (LASA).
Our LAN has achieved SOTA results on both RAW based (SID) and RGB based (LOL) low-light image enhancement datasets with a small amount of parameters.
The LASA attention mechanism proposed in this paper enables convolution operations to have the ability to establish long-range dependencies through refining feature maps, thereby improving the performance of convolution neural networks.
A large number of comparative experiments and ablation experiments have verified the effectiveness of the proposed method.
In future work, we will further explore the performance of this method on other image restoration tasks.

%% The file named.bst is a bibliography style file for BibTeX 0.99c
\bibliographystyle{named}
\bibliography{ijcai22}

\end{document}